\documentclass[conference]{IEEEtran}
\IEEEoverridecommandlockouts
\usepackage{cite}
\usepackage{amsmath,amssymb,amsfonts}
\usepackage{algorithmic}
\usepackage{graphicx}
\usepackage{textcomp}
\usepackage{xcolor}

\usepackage{caption}
\usepackage{multirow}

\def\BibTeX{{\rm B\kern-.05em{\sc i\kern-.025em b}\kern-.08em
    T\kern-.1667em\lower.7ex\hbox{E}\kern-.125emX}}
\begin{document}

\title{A Weakly Supervised and Globally Explainable Learning Framework for Brain Tumor Segmentation\\
\thanks{*Yi Pan and Yunpeng Cai are the corresponding authors.}
}


\author{\IEEEauthorblockA{Ruitao Xie\textsuperscript{1, 2}, Limai Jiang\textsuperscript{1, 2}, Xiaoxi He\textsuperscript{1}, Yi Pan\textsuperscript{1, 3*}, Yunpeng Cai\textsuperscript{1*}}
	\IEEEauthorblockA{\textit{\textsuperscript{1}Shenzhen Institute of Advanced Technology, Chinese Academy of Sciences, Shenzhen 518055, China} \\
		\textit{\textsuperscript{2}University of Chinese Academy of Sciences, Beijing 100049, China}\\
		\textit{\textsuperscript{3}Shenzhen Key Laboratory of Intelligent Bioinformatics, Shenzhen 518055, China
		}
		\\
		\{rt.xie, lm.jiang2, xjy109204, yi.pan, yp.cai\}@siat.ac.cn}}

\maketitle

\begin{abstract}
Machine-based brain tumor segmentation can help doctors make better diagnoses. However, the complex structure of brain tumors and expensive pixel-level annotations present challenges for automatic tumor segmentation. In this paper, we propose a counterfactual generation framework that not only achieves exceptional brain tumor segmentation performance without the need for pixel-level annotations, but also provides explainability. Our framework effectively separates class-related features from class-unrelated features of the samples, and generate new samples that preserve identity features while altering class attributes by embedding different class-related features. We perform topological data analysis on the extracted class-related features and obtain a globally explainable manifold, and for each abnormal sample to be segmented, a meaningful normal sample could be effectively generated with the guidance of the rule-based paths designed within the manifold for comparison for identifying the tumor regions. We evaluate our proposed method on two datasets, which demonstrates superior performance of brain tumor segmentation. The code is available at https://github.com/xrt11/tumor-segmentation.
\end{abstract}


\section{Introduction}
Brain tumors refer to an abnormal growth or mass of cells in the brain, which can be either cancerous (malignant) or non-cancerous (benign). Brain tumors can cause various neurological symptoms, such as headaches, cognitive impairments, intracranial hemorrhage and more. Diagnosing brain tumors entirely by humans faces challenges due to irregular shape, as well as the poor contrast and blurred boundaries of tumor tissues, besides, the existence of a large number of patients puts enormous pressure on the already scarce medical resources. Therefore, accurate segmentation of brain tumors using machines can greatly assist doctors in diagnosing and providing improved treatments for patients. Numerous studies have investigated the segmentation of brain tumors using MRI images, including traditional machine learning algorithms, such as random forest [1, 2], support vector machine [3, 4], and deep learning algorithms like fully convolutional networks [5], cascaded CNNs [6], dilated/atrous convolutions [7], top-down/bottom-up networks [8, 9]. However, all these methods require pixel-level annotations, which is both expensive and challenging to implement in clinical settings, especially for complex diseases like brain tumors. There are some brain tumor segmentation methods without the relying of pixel-level annotations, including thresholding-based methods [10, 11] and clustering-based segmentation methods [12, 13]. However, these traditional approaches have limited accuracy and are prone to interference due to weak ability in complex features extraction.

In recent years, some deep learning based methods that only rely on image-level class labels and do not require pixel-level annotation for training for objects localization/segmentation have been proposed, which are known as weakly supervised object localization (WSOL) [14, 15]. The most commonly used technique for WSOL is the class activation map (CAM) [16], which tends to underestimate object regions as it only identifies the most discriminative parts of an object, rather than the entire object area, resulting in poor localization performance. To address this issue, Gao et al. proposed the token semantic coupled attention map (TS-CAM) approach, which splits the image into multiple patch tokens and makes these tokens aware of object categories for more accurate object localization [17]. Considering that splitting operations can lead to loss of spatial coherence of objects, Bai et al. introduced the Spatial Calibration Module (SCM) to produce activation maps with sharper boundaries [18]. There are other weakly supervised methods that have also yielded promising results recently [19, 20]. These methods improve the accuracy of object localization or segmentation compared to CAM. However, due to the complex interactions between features and context, local gradients in these methods are susceptible to local traps, resulting in biased [21], unstable [22], inaccurate or even misleading [23, 24] localization results. Additionally, persistent criticisms [25, 26] that highlight feature associations provide limited semantic knowledge about the decision rules. Besides, these methods lack global explainability, making their widespread application in medical scenarios more challenging. 

In this paper, we present a new weakly supervised learning method that effectively addresses the challenges mentioned above. Specifically, we propose a class association embedding (CAE) framework, which includes a symmetrical cyclic generation adversarial network and a paired random shuffle training (PRST) scheme. This framework is designed to embed class associations and transform the class style of brain images while preserving individual identity features. The network consists of one encoder with two sub-modules, one decoder, and one multi-class discriminator. One module in the encoder is responsible for extracting class-related features, generating class-style (CS) codes within a unified domain. The other module focuses on extracting class-unrelated features, acquiring individual-style (IS) codes. We perform topological data analysis on extracted class-related features across the entire dataset and obtain a globally explainable manifold. For each exemplar to be segmented, we design a shortest disease-to-normal class transfer path between this exemplar and the reference counter-exemplar in this manifold. Then we acquire meaningful CS codes along this path and combine them with the IS code of the exemplar, for synthesizing new samples using the decoder in the CAE framework. By comparing the synthesized normal sample with the original abnormal exemplar, we can found and segment the tumor regions. With the assistance of the global and explainable knowledge within the class-related manifold, we can achieve more accurate and robust brain tumor segmentation results compared to other weakly supervised methods.

\section{Methods}
\subsection{Overall Framework for Brain Tumor Segmentation}
\begin{figure*}
	\centerline{\includegraphics[scale=0.07]{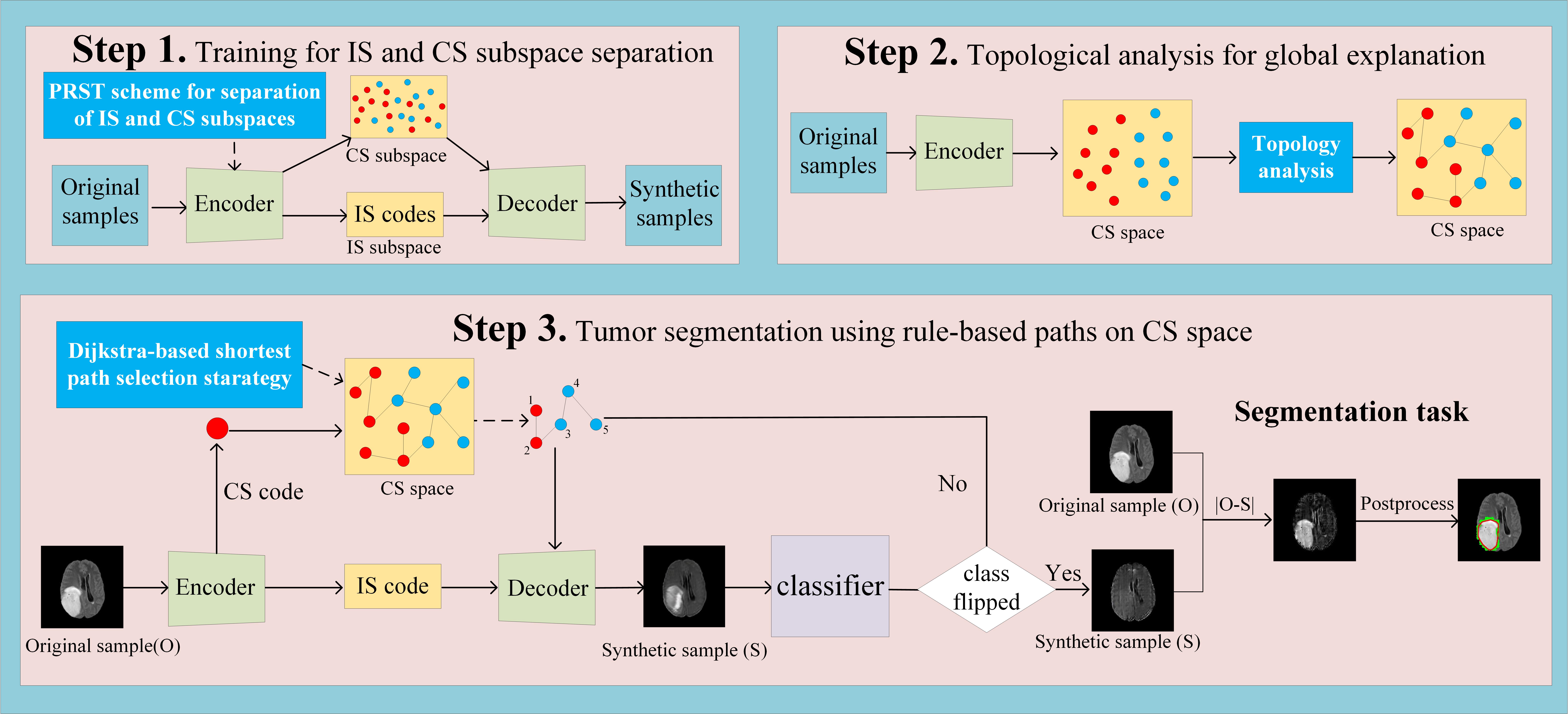}}
	\captionsetup{font={footnotesize}}
	\caption{Overall framework of brain tumor segmentation.}
	\label{fig1}
\end{figure*}

Our proposed framework for brain tumor segmentation consists of three stages, as shown in Fig. \ref{fig1}. In the first stage, we develop a symmetrical and cyclic generative adversarial network to extract class-related features and create a unified manifold. In the second stage, we utilize the trained encoder from the first stage to extract class-style (CS) codes of all samples from the dataset. We then perform topological data analysis on these extracted CS codes to generate a globally explainable topological graph, which represents the interrelationships and dependencies among the CS codes.

In the third stage, firstly, for one exemplar to be segmented, we choose a goal counter exemplar (reference) and use the trained encoder from the first stage to extract their respective CS codes. Then we select nodes on the topology graph where these two CS codes are involved, as the original and goal nodes. By performing matrix operations and the well-known Dijkstra algorithm, we determine the shortest path between these two nodes. We combine the IS code of the segmented exemplar with the center vectors of CS codes involved in each node along this path, to synthesize a series of samples. This synthesis continues until the classifier predicts the synthetic sample as the flipped class. By comparing the original exemplar with the generated counter exemplar, we are able to effectively locate and segment the brain tumor. 

\subsection{Class Association Embedding for Counterfactual Generation}

\begin{figure*}
	\centerline{\includegraphics[scale=0.11]{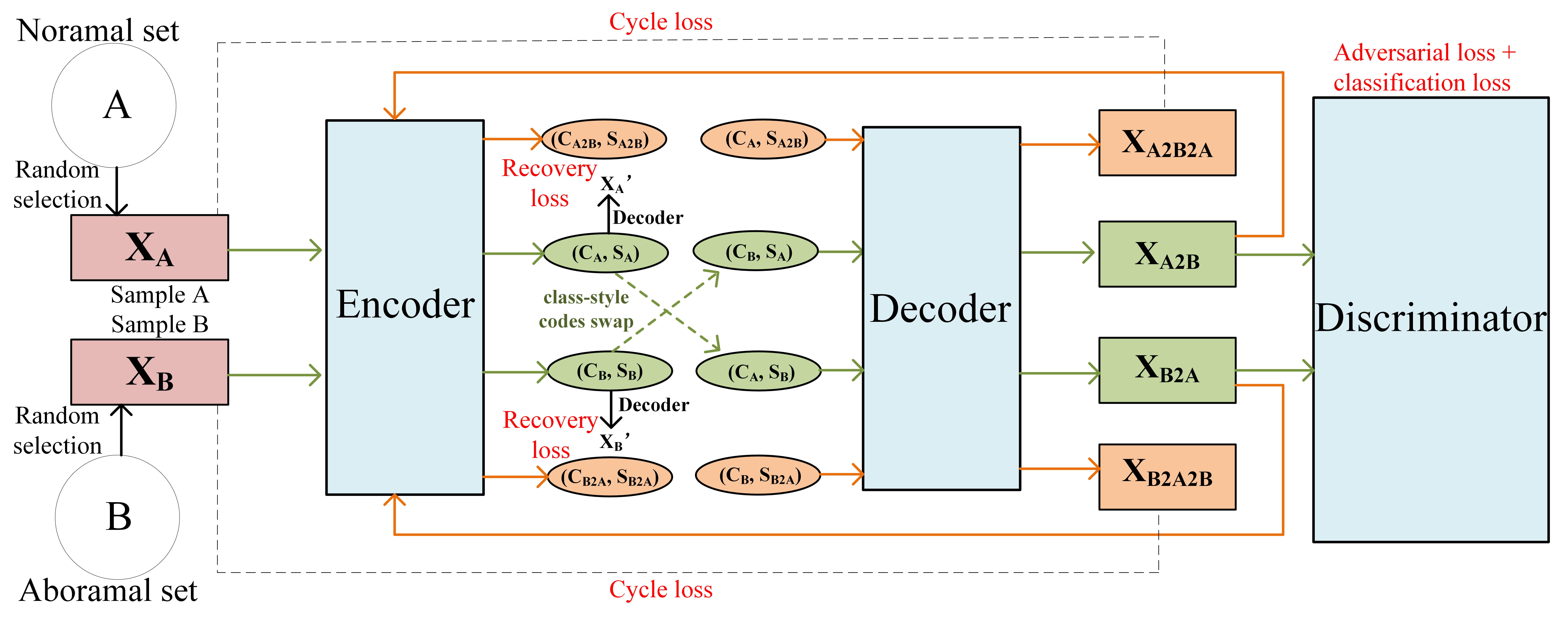}}
	\captionsetup{font={footnotesize}}
	\caption{Symmetric cyclic adversarial network with paired random shuffle training (PRST) scheme, where $X$ represents the sample, while $C$ and $S$ refer to class-style and individual-style codes respectively.}
	\label{CAE}
\end{figure*}

We develop a class association embedding (CAE) framework for counterfactual generation to achieve brain tumor segmentation. As shown in Fig. \ref{CAE}, this framework consists of a symmetrical cyclic generation adversarial network and a paired random shuffle training (PRST) scheme. The network consists of an encoder, a decoder, and a discriminator. The encoder incorporates two sub-modules (denoted by $E_{c}$ and $E_{s}$) responsible for encoding the class-related information and the individual identity information of the images separately. This results in the generation of class-style (CS) codes (denoted by $C$) and individual-style (IS) codes (denoted by $S$), respectively. The decoder (denoted by $G$) takes a combination of the IS and the CS codes as input, and generates a new sample. The discriminator (denoted by $D$) is designed for guiding the generated images to be real and with accurate class attributes. In order to preserve the class-unrelated/identity information of the original image whose IS code is used, while present the class-related features of the reference image whose CS code is employed, it is necessary for the framework to separate the class-related information from the class-unrelated information as much as possible. To achieve this goal, we propose a paired random shuffle training (PRST) scheme with a series of loss functions, where a building-block coherency feature extraction (BBCFE) method is appiled. 

During each iteration of the training process, we randomly select a normal brain image and an abnormal brain image to pair to input the network. By swapping different CS codes, we aim to generate new images that preserve identity features but achieve successful class transferring. If the classes of the generated images do not match expectations, it will be punished by the discriminator. Therefore, the CS codes are learned to contain class-related information, and due to the low dimensionality of the CS codes, the IS codes are learned to contain class-unrelated/identity information in order to achieve reconstruction. After multiple iterations of training, these two kinds of information could be separated successfully. Detailed networks design and training process are available in the previous work [27]. By combining the CS code of a normal brain image with the IS code of the original abnormal image, a new tumor-free normal brain image could be generated. By comparing the generated image with the original abnormal image, the tumor regions can be identified.

\subsection{Topology Analysis of CS Codes for Exploring and Explaining Global Rule}
By utilizing the CAE framework, class-related features are extracted from the entire dataset, resulting in a large number of 8-dimensional CS codes. To visually explore and explain global rules within the learned class-related manifold, we further perform topological data analysis (TDA) [28] on these CS codes, which presents significant connections between samples and potential structural patterns of the data through a topological representation. Specifically, we perform dimensionality reduction on these CS codes using t-SNE and divided them into multiple overlapping covers. Within each cover, Density-Based Spatial Clustering of Applications with Noise (DBSCAN) [29] is used for clustering analysis to identify samples belonging to the same class, which are represented by a single topological node. Connected lines are created between two nodes that share samples. After that, we obtain a topological graph with numerous nodes and connecting lines, allowing us to analyze sample relations, explore and visually explain global rules within the learned manifold.

\subsection{Designing Rule-based Paths on Topology Graph for Brain Tumor Segmentation}
By employing the CAE framework and topological analysis, a topological graph that encompasses global knowledge and unveils class transfer patterns could be generated, and we can design rule-based paths for class transfer, thereby facilitating the generation of meaningful counterfactuals, which enables efficient and explainable brain tumor segmentation. Specifically, for each sample to be segmented, we first select its nearest node (with the lowest Euclidean distance between the CS code of this sample and the center vector of CS codes involved in the node) as the initial node. We then randomly select another opposite-class node as the goal node and establish the shortest path between the initial and goal nodes based on the Dijkstra algorithm. To represent the topology graph, we calculate and derive an adjacency matrix initially. The entries in this matrix correspond to the Euclidean distances between the center vectors (mean values of CS codes involved in the node) of the connected nodes (i.e., node $i$ and node $j$). For node pairs without a direct connection, the values are assigned as infinity. By utilizing this distance matrix and performing the Dijkstra operation, we can determine the shortest paths between every pair of nodes. We can efficiently obtain meaningful CS codes (center vectors of the nodes) along this path, and combine them with the IS code of the sample to be segmented to generate reasonable counterfactual examples for comparison. This approach produces explainable and accurate segmentation results that conform to global knowledge rules.

\section{Experiment}
\subsection{Datasets and Implementation Details}
We utilized two datasets, BraTS2020 [30] and BraTS2021 [31], for our experiments. For the BraTS2020 dataset, a total of 1,298 brain images were used in our experiments. Among these, 1,005 images contained tumors, while 293 images were normal. We randomly selected 704 abnormal brain images and 206 normal images for the training set, and the remaining images formed the test set. For the BraTS2021 dataset, we randomly selected 3,828 abnormal brain images and 710 normal images for the training set. And the test set comprised 1,623 abnormal images and 302 normal images. 

In our experimental setup, we resize all input images to a size of 256 $\times$ 256 pixels. During training, we randomly apply horizontal flipping to the input images with a probability of 0.5. To update the network parameters, we utilize the Adam optimizer [32] with an initial learning rate and weight decay set to 1e-4. 

\subsection{Evaluation Metrics}
In order to quantitatively evaluate the effectiveness of the method we have proposed. We introduced the Intersection over Union (IOU) and Dice Similarity Coefficient (DICE) metrics, which were widely employed in computer vision and image segmentation tasks to quantify the similarity between two segmented regions or images. IOU assesses the overlap between two sets of pixels in an image, computed by dividing the area of intersection by the area of the union of the two sets of pixels. DICE, on the other hand, is a metric that evaluates the similarity between two segmented regions or images based on the boundary overlap. It is calculated by dividing the sum of the squared similarity coefficients of the corresponding pixels in the two regions by the sum of the squared dissimilarity coefficients of the corresponding pixels. Both IOU and DICE score ranges from 0 to 1, with 1 indicating a perfect match between the two regions or images, and 0 indicating no similarity whatsoever. We calculated the IOU score and the DICE score between the predicted mask and the ground truth mask for each test image, and took the mean IOU score and the mean DICE score as our final results for performance evaluation, as expressed in eq.\ref{con:loss5} and eq.\ref{con:loss6}, where $I_{p, k}$ represents the predicted mask (height: $H$, width: $W$, consists of pixels with value 0 or 1, 0 means negative and 1 means positive) for the k-th image ($N$ in total), and $I_{p, k}$ $(i, j)$ refers to the pixel value of $I_{p, k}$ at the position index $(i, j)$, while $I_{g, k}$ denotes the corresponding ground truth mask (consists of pixels with value 0 or 1, 0 means negative and 1 means positive).

\begin{equation}
	\small
	S (I_{p, k}) = \sum_{i=0}^{i=H-1}\sum_{j=0}^{j=W-1}I_{p, k}(i, j)
	\label{con:loss1}
\end{equation}
\begin{equation}
	\small
	S (I_{g, k}) = \sum_{i=0}^{i=H-1}\sum_{j=0}^{j=W-1}I_{g, k}(i, j)
	\label{con:loss2}
\end{equation}
\begin{equation}
	\small
	S (I_{p, k} \cap I_{g, k}) = \sum_{i=0}^{i=H-1}\sum_{j=0}^{j=W-1}[I_{p, k}(i, j) × I_{g, k}(i, j)]
	\label{con:loss3}
\end{equation}
\begin{equation}
	\small
	S (I_{p, k} \cup I_{g, k}) = S (I_{p, k})+S (I_{g, k})-S (I_{p, k} \cap I_{g, k})
	\label{con:loss4}
\end{equation}
\begin{equation}
	\small
	mIOU  = (1/N) \times \sum_{k=1}^{k=N}S (I_{p, k} \cap I_{g, k})/S (I_{p, k} \cup I_{g, k})
	\label{con:loss5}
\end{equation}
\begin{equation}
	\small
	mDICE  = (1/N) \times \sum_{k=1}^{k=N}[2 \times S (I_{p, k} \cap I_{g, k})]/[S (I_{p, k}) + S (I_{g, k})]
	\label{con:loss6}
\end{equation}

%
%

\subsection{Results}
\textbf{Segmentation Results.} We conducted experiments to validate the effectiveness of our proposed method for brain tumor segmentation, as described in Section 2. By comparing the original samples with the synthetic samples obtained by embedding meaningful CS codes along the rule-based path within the class-related manifold, difference maps were acquired and then post-processed to generate the segmentation results. To demonstrate the superiority of our method in the brain tumor segmentation task, we compared our proposed segmentation method with other existing weakly supervised learning algorithms (TSCAM [17], ICAM [33], SCAM [18], Bridging [19] and LAGAN [20]). We used IOU and DICE metrics for quantitative comparison. The results in Table \ref{seg-results} indicate higher mean IOU and DICE values on both datasets using our proposed method, providing evidence that our approach outperforms others in the brain tumor segmentation task. 

\begin{figure}
	
	\centerline{\includegraphics[scale=0.205]{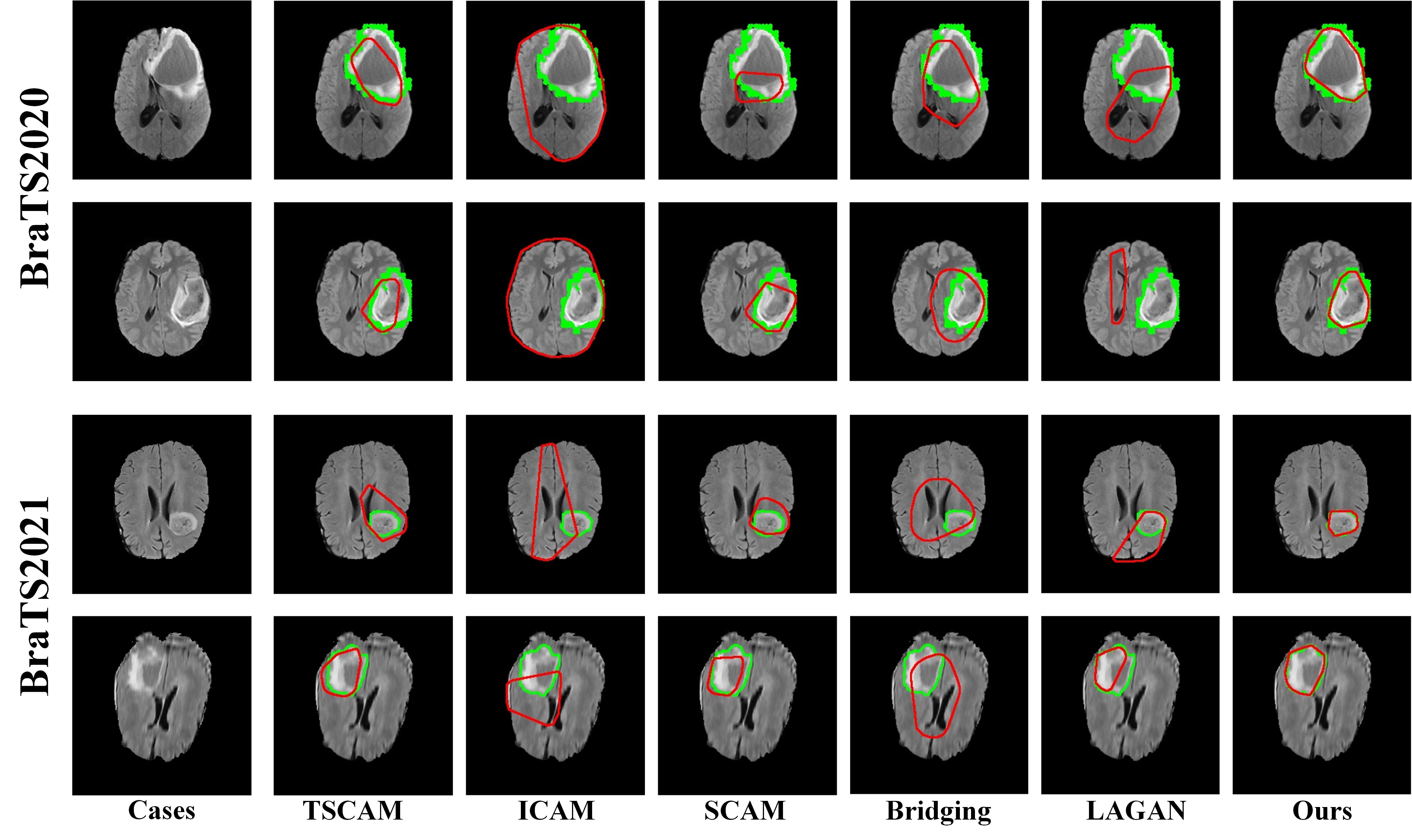}}
	\captionsetup{font={footnotesize}}
	\caption{Segmentation cases using different algorithms. The regions surrounded by the green lines are the groundtruth, while the regions surrounded by the red lines are the predicted results.}
	\label{seg-cases-compare}
\end{figure}

In order to more intuitively demonstrate the superiority of our algorithm. We showed several segmented cases in Fig. \ref{seg-cases-compare}. It could be well seen that better segmentation results achieved with our approach, further illustrating the superiority of our algorithm.

\begin{table}
	\caption{\centering{IOU and DICE results on BraTS2020 and BraTS2021 test sets using different algorithms}} 
	\label{seg-results}
	\centering
	\begin{tabular}{l|cc|cc}
		\hline
		\multirow{2}*{Methods}    &\multicolumn{2}{c|}{BraTS2020}  &\multicolumn{2}{c}{BraTS2021}  \\
		\cline{2-5}
		~  &IOU  &DICE   &IOU  &DICE \\
		\hline
		TSCAM   &0.5810 &0.6972 &0.5497 &0.6540  \\
		ICAM   &0.5729 &0.7138 &0.2814 &0.4002  \\
		SCAM   &0.5378 &0.6657 &0.5375 &0.6495  \\
		Bridging   &0.4549 &0.6180 &0.2158 &0.3470  \\
		LAGAN  &0.4214 &0.5556 &0.4220 &0.5420  \\
		\hline
		CAE (ours)  &\textbf{0.6373} &\textbf{0.7585} &\textbf{0.5791} &\textbf{0.6915}  \\
		\hline
	\end{tabular}
\end{table}

		%
		%

\begin{figure}
	\centerline{\includegraphics[scale=0.24]{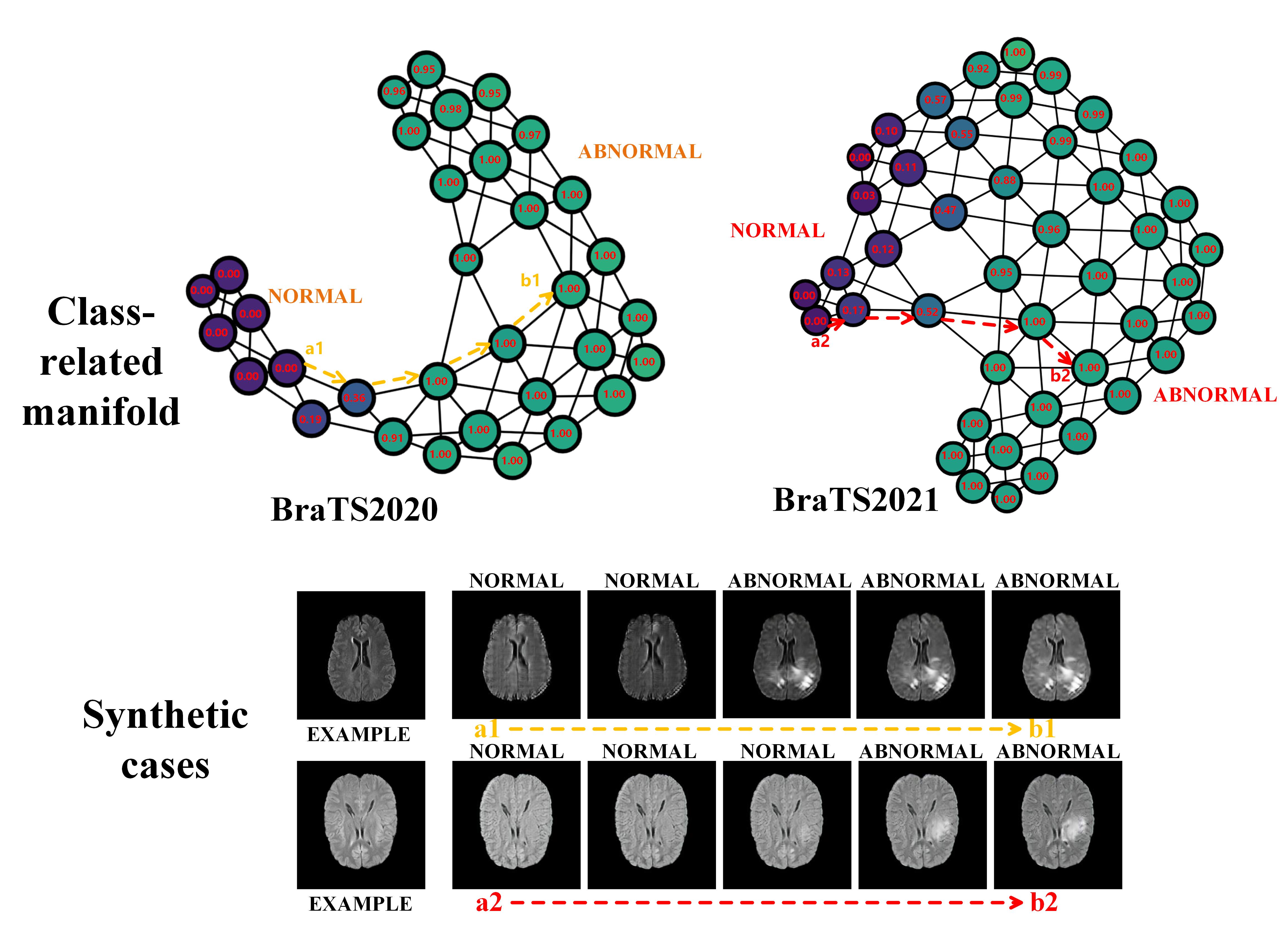}}
	\captionsetup{font={footnotesize}}
	\caption{The subgraph above is the results of topology analysis (topology graphs) of learned class-related manifolds on BraTS2020 and BraTS2021 datasets, while the subgraph below is some synthetic cases obtained based on the defined paths within the manifolds. In the topology graphs, values with red font inside the nodes refer to the ratios of the abnormal cases involved in these nodes. In the synthetic cases, the IS codes of the EXAMPLE are extracted for combinations with the center vectors of the CS codes involved in each node along the defined paths. The predicted classes by the external classifiers are presented above the synthetic cases.}
	\label{latent-topology-analysis-generate}
\end{figure}

\textbf{Explainability Results.}
In addition to achieving improved segmentation results, our algorithm also possesses strong explainability. We performed topological data analysis on the CS codes from the test sets. The results of the topology analysis are shown in Fig. \ref{latent-topology-analysis-generate}, providing insights into the characteristics and structure of the learned manifold. The visual representations obtained through topology analysis clearly show a separation between normal and abnormal cases in the learned class-related manifold. It is worth noting that normal cases tend to cluster towards the left margin, while the proportion of abnormal cases gradually increases as the paths extend towards the right on the topology graph. This alignment with the progression of disease development provides valuable and explainable insights. We randomly chose two normal examples and designed two guided paths as shown in Fig. \ref{latent-topology-analysis-generate}, to generate a series of new images, which show a noticeable trend of tumor development and the emergence of more pronounced disease-related characteristics along the defined paths. This further confirms the successful acquisition of rule-based and explainable knowledge within our class-related manifold. By leveraging this topology graph, we design class transfer paths that adhere to medical knowledge, enabling us to acquire meaningful CS codes for embedding and counterfactual generation, and resulting in a more explainable segmentation results.


\subsection{Ablation Studies}
\textbf{Effectiveness of Topology Analysis for Brain Tumor Segmentation.}
Topological analysis offers numerous connections that can elucidate the relationship between samples/classes, enabling the discovery of efficient, accurate and rule-based class transfer paths for meaningful counterfactual generation for accurate segmentation. To assess the efficacy of topological analysis for brain tumor segmentation, we conducted an ablation study. The experiment results were presented in Table \ref{abstudy}, where $wo/TDA$ and $w/TDA$ refers to the experiments without and with topological data analysis respectively. For $wo/TDA$, owing to the absence of topological connection maps, we could only establish a straight line between the original detected samples and the target reference samples. We sampled CS codes at regular intervals of 0.1$d$ (where $d$ represents the Euclidean distance between the CS codes of the original and reference samples) from this linear path for embedding and counterfactual generation for brain tumor segmentation. The experimental results reveal that, with the assistance of the topological map, superior segmentation results can be achieved. This is attributed to the topological map's ability to facilitate the identification of accurate, shortest, and rule-based class transfer paths. 

\textbf{Effectiveness of Path Design for Brain Tumor Segmentation.}
Designing rule-based class transfer paths within the global topological graph for counterfactual generation could improve explainability for brain tumor segmentation. Besides, along these paths, we can identify class transformation edges and generate class-flipping counterfactual sample that are more akin to the original ones. This is advantageous for reducing unnecessary information and interference, and achieving more accurate segmentation results by comparison. We conducted ablation study to confirm the effectiveness of paths design, and the results are presented in Table \ref{abstudy}, where $wo/paths$ refers to the experiment that counterfactual samples were directly obtained using the CS codes from the target reference images for comparison, while $w/paths$ refers to the experiment that counterfactual samples were generated along designed paths as mentioned above. The findings indicate that designing guided paths is beneficial for brain tumor segmentation.


\begin{table}
	\begin{center}
		\caption{\centering{IOU and DICE results on BraTS2020 and BraTS2021 test sets}} \label{abstudy}
		\begin{tabular}{l|cc|cc}
			\hline
			\multirow{2}*{Methods}    &\multicolumn{2}{c|}{BraTS2020}  &\multicolumn{2}{c}{BraTS2021}  \\
			\cline{2-5}
			~    &IOU  &DICE  &IOU  &DICE  \\
			\hline
			$wo/TDA$   &0.4516 &0.5752 &0.5372 &0.6482 \\
			$w/TDA$  &0.6373 &0.7585 &0.5791 &0.6915 \\
			\hline
			$wo/paths$  &0.4050 &0.5253 &0.5160 &0.6245 \\
			$w/paths$ &0.6373 &0.7585 &0.5791 &0.6915 \\
			\hline
		\end{tabular}
	\end{center}
\end{table}

\section{Conclusion}
Automatic segmentation of brain tumors is a crucial and exceptionally challenging task. Current approaches for brain tumor segmentation often rely on expensive pixel-level annotations, lack explainability and perform poorly. In this study, we propose a weakly supervised and explainable learning framework to address these common issues. Our proposed framework successfully learns a globally explainable class-related manifold, and we can perform counterfactual generation that adheres to medical rules by actively designing rule-based class transfer paths within this manifold for explainable segmentation, without the reliance of pixel-level annotations. Besides, with the assistance of the global knowledge learned in the class-related manifold, local traps could be well avoided and more accurate segmentation results could be achieved using our proposed method compared to other existing algorithms.

\section*{Acknowledgment}

This work was supported by the Strategic Priority Research Program of Chinese Academy of Sciences (grant no. XDB38050100), Shenzhen Science and Technology Program under grant no. KQTD20200820113106007, Shenzhen Key Laboratory of Intelligent Bioinformatics (ZDSYS20220422103800001), the National Natural Science Foundation of China under grant No. U22A2041.

%
%
%
%



\begin{thebibliography}{00}


\bibitem{b1} Subhranil Koley, Anup K Sadhu, Pabitra Mitra, Basabi Chakraborty,
and Chandan Chakraborty, “Delineation and diagnosis of brain tumors
from post contrast t1-weighted mr images using rough granular
computing and random forest,” Applied Soft Computing, vol. 41, pp.
453–465, 2016.
\bibitem{b2} Mohammadreza Soltaninejad, Lei Zhang, Tryphon Lambrou, Guang
Yang, Nigel Allinson, and Xujiong Ye, “Mri brain tumor segmentation
and patient survival prediction using random forests and fully
convolutional networks,” in Brainlesion: Glioma, Multiple Sclerosis,
Stroke and Traumatic Brain Injuries: Third International Workshop,
BrainLes 2017, Held in Conjunction with MICCAI 2017, Quebec City,
QC, Canada, September 14, 2017, Revised Selected Papers 3. Springer,
2018, pp. 204–215.
\bibitem{b3} T Sathies Kumar, K Rashmi, Sreevidhya Ramadoss, LK Sandhya, and
TJ Sangeetha, “Brain tumor detection using svm classifier,” in 2017
Third International Conference on Sensing, Signal Processing and Security
(ICSSS). IEEE, 2017, pp. 318–323.
\bibitem{b4} Ahmed Kharrat, Karim Gasmi, M Ben Messaoud, Nac´era Benamrane,
and Mohamed Abid, “A hybrid approach for automatic classification
of brain mri using genetic algorithm and support vector machine,”
Leonardo journal of sciences, vol. 17, no. 1, pp. 71–82, 2010.
\bibitem{b5} Jindong Sun, Yanjun Peng, Yanfei Guo, and Dapeng Li, “Segmentation
of the multimodal brain tumor image used the multi-pathway architecture
method based on 3d fcn,” Neurocomputing, vol. 423, pp. 34–45,
2021.
\bibitem{b6} Ramin Ranjbarzadeh, Abbas Bagherian Kasgari, Saeid Jafarzadeh
Ghoushchi, Shokofeh Anari, Maryam Naseri, and Malika
Bendechache, “Brain tumor segmentation based on deep learning and
an attention mechanism using mri multi-modalities brain images,”
Scientific Reports, vol. 11, no. 1, pp. 10930, 2021.
\bibitem{b7} Daniel E Cahall, Ghulam Rasool, Nidhal C Bouaynaya, and Hassan M
Fathallah-Shaykh, “Dilated inception u-net (diu-net) for brain tumor
segmentation,” arXiv preprint arXiv:2108.06772, 2021.
\bibitem{b8} Xi Guan, Guang Yang, Jianming Ye, Weiji Yang, Xiaomei Xu, Weiwei
Jiang, and Xiaobo Lai, “3d agse-vnet: an automatic brain tumor mri
data segmentation framework,” BMC medical imaging, vol. 22, no. 1,
pp. 1–18, 2022.
\bibitem{b9} Yongchao Jiang, Mingquan Ye, Peipei Wang, Daobin Huang, and Xiaojie
Lu, “Mrf-iunet: A multiresolution fusion brain tumor segmentation
network based on improved inception u-net,” Computational and
Mathematical Methods in Medicine, vol. 2022, 2022.
\bibitem{b10} Myat Thet Nyo, F Mebarek-Oudina, Su Su Hlaing, and Nadeem A
Khan, “Otsu’s thresholding technique for mri image brain tumor segmentation,”
Multimedia tools and applications, vol. 81, no. 30, pp.
43837–43849, 2022.
\bibitem{b11} Rasha Khilkhal and Mustafa Ismael, “Brain tumor segmentation utilizing
thresholding and k-means clustering,” in 2022 Muthanna International
Conference on Engineering Science and Technology (MICEST).
IEEE, 2022, pp. 43–48.
\bibitem{b12} Md Shahariar Alam, Md Mahbubur Rahman, Mohammad Amazad
Hossain, Md Khairul Islam, Kazi Mowdud Ahmed, Khandaker Takdir
Ahmed, Bikash Chandra Singh, and Md Sipon Miah, “Automatic human
brain tumor detection in mri image using template-based k means
and improved fuzzy c means clustering algorithm,” Big Data and Cognitive
Computing, vol. 3, no. 2, pp. 27, 2019.
\bibitem{b13} Robert Setyawan, Mustofa Alisahid Almahfud, Christy Atika Sari,
Eko Hari Rachmawanto, et al., “Mri image segmentation using morphological
enhancement and noise removal based on fuzzy c-means,”
in 2018 5th international conference on information technology, computer,
and electrical engineering (ICITACEE). IEEE, 2018, pp. 99–104.
\bibitem{b14} Junsuk Choe and Hyunjung Shim, “Attention-based dropout layer
for weakly supervised object localization,” in Proceedings of the
IEEE/CVF Conference on Computer Vision and Pattern Recognition,
2019, pp. 2219–2228.
\bibitem{b15} Xingjia Pan, Yingguo Gao, Zhiwen Lin, Fan Tang, Weiming Dong,
Haolei Yuan, Feiyue Huang, and Changsheng Xu, “Unveiling the potential
of structure preserving for weakly supervised object localization,”
in Proceedings of the IEEE/CVF Conference on Computer Vision
and Pattern Recognition, 2021, pp. 11642–11651.

\bibitem{b16} Bolei Zhou, Aditya Khosla, Agata Lapedriza, Aude Oliva, and Antonio
Torralba, “Learning deep features for discriminative localization,” in
Proceedings of the IEEE conference on computer vision and pattern
recognition, 2016, pp. 2921–2929.
\bibitem{b17} Yuan Yao, Fang Wan, Wei Gao, Xingjia Pan, Zhiliang Peng, Qi Tian,
and Qixiang Ye, “Ts-cam: Token semantic coupled attention map for
weakly supervised object localization,” IEEE Transactions on Neural
Networks and Learning Systems, 2022.
\bibitem{b18} Haotian Bai, Ruimao Zhang, Jiong Wang, and Xiang Wan, “Weakly
supervised object localization via transformer with implicit spatial calibration,”
in European Conference on Computer Vision. Springer, 2022,
pp. 612–628.
\bibitem{b19} Eunji Kim, Siwon Kim, Jungbeom Lee, Hyunwoo Kim, and Sungroh
Yoon, “Bridging the gap between classification and localization
for weakly supervised object localization,” in Proceedings of the
IEEE/CVF Conference on Computer Vision and Pattern Recognition,
2022, pp. 14258–14267.
\bibitem{b20} Yuhui Tao, Xiao Ma, Yizhe Zhang, Kun Huang, Zexuan Ji, Wen Fan,
Songtao Yuan, and Qiang Chen, “Lagan: Lesion-aware generative
adversarial networks for edema area segmentation in sd-oct images,”
IEEE Journal of Biomedical and Health Informatics, 2023.
\bibitem{b21} Avanti Shrikumar, Peyton Greenside, and Anshul Kundaje, “Learning
important features through propagating activation differences,” in
International conference on machine learning. PMLR, 2017, pp. 3145–
3153.
\bibitem{b22} Julius Adebayo, Justin Gilmer, Michael Muelly, Ian Goodfellow,
Moritz Hardt, and Been Kim, “Sanity checks for saliency maps,” Advances
in neural information processing systems, vol. 31, 2018.
\bibitem{b23} Lieyang Chen, Anthony Cruz, Steven Ramsey, Callum J Dickson,
Jose S Duca, Viktor Hornak, David R Koes, and Tom Kurtzman, “Hidden
bias in the dud-e dataset leads to misleading performance of deep
learning in structure-based virtual screening,” PloS one, vol. 14, no. 8,
pp. e0220113, 2019.
\bibitem{b24} Samek, Wojciech, et al., eds. “Explainable AI: interpreting, explaining
and visualizing deep learning,” vol. 11700, Springer Nature,
2019.
\bibitem{b25} Marzyeh Ghassemi, Luke Oakden-Rayner, and Andrew L Beam, “The
false hope of current approaches to explainable artificial intelligence in
health care,” The Lancet Digital Health, vol. 3, no. 11, pp. e745–e750,
2021.
\bibitem{b26} Cynthia Rudin, “Stop explaining black box machine learning models
for high stakes decisions and use interpretable models instead,” Nature
machine intelligence, vol. 1, no. 5, pp. 206–215, 2019.
\bibitem{b27} Ruitao Xie, Jingbang Chen, Limai Jiang, Rui Xiao, Yi Pan, Yunpeng Cai, ``Accurate Explanation Model for Image Classifiers using Class Association Embedding,'' 2024 IEEE 40th International Conference on Data Engineering (ICDE). IEEE, 2024. Accepted.
\bibitem{b28} Gunnar Carlsson, “Topology and data,” Bulletin of the American Mathematical
Society, vol. 46, no. 2, pp. 255–308, 2009.
\bibitem{b29} Martin Ester, Hans-Peter Kriegel, J¨org Sander, Xiaowei Xu, et al.,
“A density-based algorithm for discovering clusters in large spatial
databases with noise,” in kdd, 1996, vol. 96, pp. 226–231.
\bibitem{b30} MICCAI, “brats2020,” https://www.med.upenn.edu/
cbica/brats2020/.
\bibitem{b31} Ujjwal Baid, Satyam Ghodasara, Suyash Mohan, Michel Bilello, Evan
Calabrese, Errol Colak, Keyvan Farahani, Jayashree Kalpathy-Cramer,
Felipe C Kitamura, Sarthak Pati, et al., “The rsna-asnr-miccai brats
2021 benchmark on brain tumor segmentation and radiogenomic classification,”
arXiv preprint arXiv:2107.02314, 2021.
\bibitem{b32} Diederik P Kingma and Jimmy Ba, “Adam: A method for stochastic
optimization,” arXiv preprint arXiv:1412.6980, 2014.
\bibitem{b33} Cher Bass, Mariana Da Silva, Carole Sudre, Logan ZJ Williams, Helena
S Sousa, Petru-Daniel Tudosiu, Fidel Alfaro-Almagro, Sean P
Fitzgibbon, Matthew F Glasser, Stephen M Smith, et al., “Icam-reg:
Interpretable classification and regression with feature attribution for
mapping neurological phenotypes in individual scans,” IEEE Transactions
on Medical Imaging, vol. 42, no. 4, pp. 959–970, 2022.















\end{thebibliography}
\end{document}